\documentclass[sigconf]{acmart}
\usepackage{lipsum}

\AtBeginDocument{%
  \providecommand\BibTeX{{%
    \normalfont B\kern-0.5em{\scshape i\kern-0.25em b}\kern-0.8em\TeX}}}

\acmConference[CONCATENATE - HRI]{}{March 13, 2023}{Stockholm, SE}

\begin{document}

\title{NetROS-5G: Enhancing Personalization through 5G Network Slicing and Edge Computing in Human-Robot Interactions}

\author{Anestis Dalgkitsis}

\email{a.dalgkitsis@iquadrat.com}
\orcid{0000-0001-6838-1768}

\affiliation{
  \institution{Research \& Development Department,\\ Iquadrat Informatica}
  \streetaddress{Carrer del Dr. Rizal, 10}
  \city{Barcelona}
  \country{Spain}
  \postcode{08006}
}

\author{Christos Verikoukis}
\email{cveri@ceid.upatras.gr}

\affiliation{
  \institution{Computer Engineering \& Informatics Department, University of Patras}
  \streetaddress{25is Martiou 7, Kato Kastritsi}
  \city{Patras}
  \country{Greece}
  \postcode{26504}
}

\affiliation{
  \institution{Iquadrat Informatica}
  \streetaddress{Carrer del Dr. Rizal, 10}
  \city{Barcelona}
  \country{Spain}
  \postcode{08006}
}

\affiliation{
  \institution{ISI/ATHENA}
  \streetaddress{Artemidos 6 & Epidavrou}
  \city{Athens}
  \country{Greece}
  \postcode{15125}
}

\begin{abstract}

Robots are increasingly being used in a variety of applications, from manufacturing and healthcare to education and customer service. However, the mobility, power, and price points of these robots often dictate that they do not have sufficient computing power on board to run modern algorithms for personalization in human-robot interaction at desired rates. This can limit the effectiveness of the interaction and limit the potential applications for these robots.
5G connectivity provides a solution to this problem by offering high data rates, bandwidth, and low latency that can facilitate robotics services. Additionally, the widespread availability of cloud computing has made it easy to access almost unlimited computing power at a low cost. Edge computing, which involves placing compute resources closer to the action, can offer even lower latency than cloud computing.
In this paper, we explore the potential of combining 5G, edge, and cloud computing to provide improved personalization in human-robot interaction. We design, develop, and demonstrate a new framework, entitled NetROS-5G, to show how the performance gained by utilizing these technologies can overcome network latency and significantly enhance personalization in robotics. 
Our results show that the integration of 5G network slicing, edge computing, and cloud computing can collectively offer a cost-efficient and superior level of personalization in a modern human-robot interaction scenario.

\end{abstract}

\keywords{5G, edge computing, cloud computing, network slicing, personalized robotics}

\maketitle

\section{Introduction}

The incorporation of advanced algorithms in robotics has led to an increasing demand for more powerful computing resources. However, the onboard computing capabilities of robots are frequently limited, and may not be able to keep pace with the rapid advancements in the field. In light of this, cloud computing has emerged as a viable solution, offering on-demand access to an abundance of computing resources. Edge computing on the other hand, provides low latency access to computing resources in close proximity to the robot, which can reduce the delay in communication between the robots and the computational resources. The emergence of 5G networks has further enabled the availability of low-latency access to edge computing resources, leading to improvements in the overall responsiveness. The integration of cloud computing, edge computing, and 5G networks has the potential to significantly expand the capabilities of robots, lower their production price and enhance the personalization capabilities and naturalness of Human-Robot Interaction (HRI), via low-latency access to even further computational resources.

5G network slicing is a feature of the 5G networks that allows for the creation of multiple virtual networks within a physical network, each with its own specific characteristics and capabilities. Edge computing, on the other hand, is a method of processing data closer to the source of the data, rather than sending it to a centralized location for processing. When combined, 5G network slicing and edge computing can provide improved human-robot interaction and personalization by allowing for the creation of dedicated virtual networks specifically designed for each use-case. These virtual networks can be optimized for low-latency, high-bandwidth communication, which is crucial for real-time interactions between humans and robots. Additionally, edge computing can be used to process sensor data from robots in real-time, allowing them to make more informed decisions and respond to human input more quickly. This can lead to more natural and intuitive interactions between humans and robots, as well as more personalized experiences.

Our key idea is that the human-robot interactions could be enhanced by the use of modern 5G and beyond networking in conjunction with edge and cloud computing. This presents a business market opportunity for cheaper robots with better performance, that offer better personalization.

To achieve this, we design and implement a framework based on the functions of 5G network slicing, edge and cloud computing to orchestrate network services with the end goal of better personalization and performance in HRI. By using off-the-shelf parts we build a 5G Stand Alone (5G-SA) network with slicing capabilities in both Radio Access Network (RAN) and computing part. We build an \textit{in situ} prototype management software, entitled NetRos-5G, 
which can seamlessly extend the capabilities of Robot Operating System 2 (ROS2) between the robot, remote cloud, and edge computing nodes.
The abundance of resources and capabilities beyond the robot can enhance the personalization capabilities and HRI.

\section{Related work}

The utilization of cloud computing in robotics, also known as cloud robotics, has been previously investigated as a viable approach for offloading tasks or increasing processing capabilities.
Several studies have proposed automated frameworks for provisioning and deploying ROS 2 nodes on public cloud computing providers to improve robot performance. Notably, FogROS, FogROS2, and FogROS G, introduced in \cite{FogROS}, \cite{FogROS2}, and \cite{FogROSG}, respectively, are among the most significant contributions. FogROS solutions are unique in that they push robot applications from the robot to the cloud, rather than remotely assisting tasks. The studies demonstrate the potential performance gains of offloading ROS 2 nodes to cloud computing. While cloud computing introduces latency, it can overcome network latency and significantly improve robot performance, depending on the task.

Recently the focus has shifted from cloud computing to the new promising edge computing that provides less computing capabilities, but with lower latency.
Giovanni Toffetti et al. in \cite{CloudNativeGPU} discuss the main advantages combining the use of a Kubernetes and real, heterogeneous, robotic hardware, the main shortcomings we encountered with networking and sharing GPUs to support deep learning workloads. 
More specifically, Peng Huang et al. in \cite{EdgeRoboticsSLAM} propose RecSLAM, a multi-robot laser Simultaneous Localization and Mapping (SLAM) built upon the Robot-edge-cloud architecture to allow for low-latency response. RecSLAM develops a hierarchical map fusion technique that directs robot raw data to edge servers for real-time fusion and then sends to the cloud for global merging. Their evaluations show RecSLAM can achieve up to 39.31\% processing latency reduction over the state-of-the-art.

The integration of edge and cloud computing has begun to emerge as a topic of interest in recent literature.
Authors in \cite{DistrRobSysEdgeCloudConnm} list several recent works that leverage the Edge-Cloud Continuum with ROS 2 and Kubernetes.

The integration of 5G technologies in robotics is a relatively novel field of study and has the potential to serve as a foundation for the development of new systems in which modern robotics can rely on. This can be achieved by combining 5G with edge and cloud computing.
The authors in \cite{validClourRob5GMEC} propose a real-time cloud robotics application by offloading robot navigation engine over to 5G Mobile Edge Computing (MEC) sever. Robot engines are containerized based on micro-service architecture and have been deployed using Kubernetes. They successfully demonstrated that mobile robots are able to avoid obstacles in real-time when the engines are remotely running in 5G MEC server.

In contrast, the proposed approach utilizes 5G technologies to selectively access a portion of the entire networking and computing infrastructure via 5G network slicing, specifically tailored for each individual task or scenario. This has shown to yield superior performance compared to other solution and enhance the HRI.

\section{Methodology \& Design}

We consider an autonomous robot robot equipped with audiovisual capabilities, including a camera, microphone, and display, to assist travelers in navigating an airport. The robot must possess the ability to comprehend spoken commands, verbally communicate, display directions, gather sensor data, and navigate while avoiding obstacles. The objective of this system is to aid travelers in reaching their designated departure gates efficiently.

\subsection{System Model Architecture}

\begin{figure}[ht]
  \includegraphics[width=\linewidth]{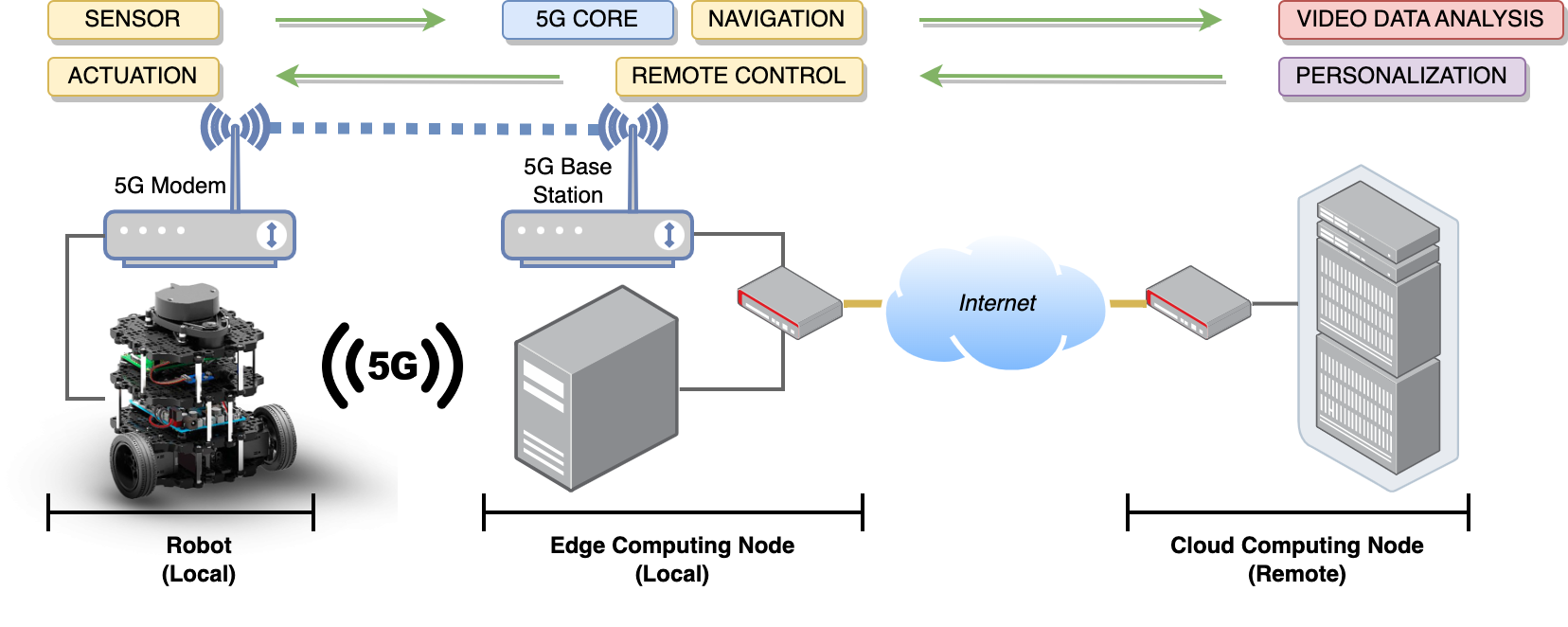}
  \caption{Architecture overview of the studied scenario. The robot is able to access the edge and cloud computing resources via the 5G network.}
  \label{fig:arch}
\end{figure}

The robot is connected to a local 5G Base Station (BS) though a 5G-enabled modem that is connected via a 10GE to the robot. A server with limited computing capabilities is connected to the BS through a switch via a 10GE, all located in close proximity to the robot in a range less than 1 km. This computing setup close to the robot is defined as a Mobile Edge Computing (MEC) node. The MEC node has access through the Internet (via the land network) to a public cloud provider that offers virtually unlimited computing resources. The studied architecture can be seen in the Fig. \ref{fig:arch}.

\subsection{Network Slice Framework}

\begin{figure*}[ht]
  \includegraphics[width=\textwidth]
  {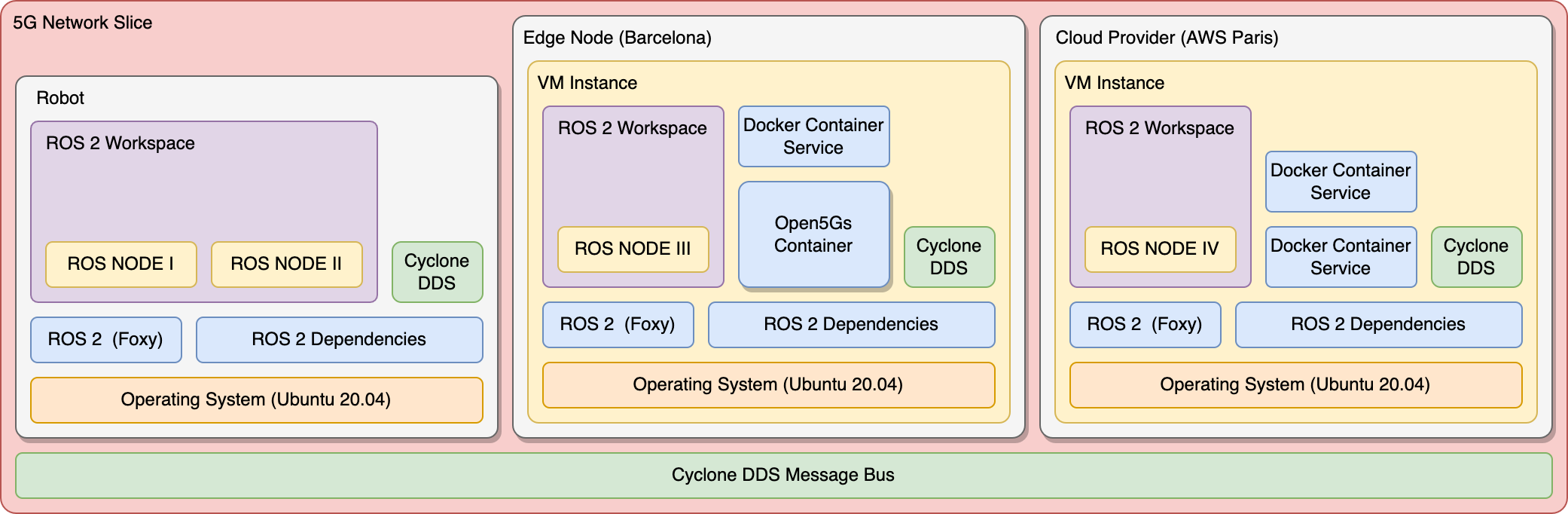}
  \caption{Overview of the network slice architecture. The network slice ensures the security and isolation of the robot functionality from the rest of the network, while also providing controlled network and computing resources.}
  \label{fig:slice}
\end{figure*}

All aforementioned entities belong in a 5G networking slice that extends from the robot, to the edge and cloud computing nodes. 
All robot ROS 2 nodes were containerized based on the modern micro-service architecture and deployed using Kubernetes \cite{KubernetesWeb} to provide scalability and load balancing functions.
A shared Cyclone DDS Messaging Bus \cite{CycloneDDSgithub} ensures fault-tolerant data exchange.
The dedicated network slice confers security and isolation from the rest of the network, thereby mitigating the risk of potential attacks.
The computational and network slice architecture can be seen in the Fig. \ref{fig:slice}.

\section{Practical Evaluation}

In this section, we present the examined scenario and results during the experimental evaluation of our case study.

\subsection{Experimental Setting}

In this study, the experimental evaluation takes place in a laboratory environment with physical 5G and computing equipment. The robot of choice, the TurtleBot3 \cite{TurtleBot3}, was simulated in an x86 mini-computer with the GAZEBO robot simulator \cite{GazeboSim}, as the delivery of the actual robot was delayed due to supply-chain disruptions. The robot was connected through an Ethernet link to a 5G-modem, branded Teltonika TRB500 \cite{TeltonikaTRB500modem}. An Amarisoft Callbox Mini \cite{AmarisoftCBminiBS} 5G BS was connected to a server cluster, used as an edge computing node. 

The Amarisoft Callbox BS is single-cell 5G NR standalone or Long-Term Evolution (LTE) cells. It is composed of a PCIe Software-Defined Radio (SDR) card, 2x2 Multiple-Input Multiple-Output (MIMO), up to 500 devices with 2 slices. The Amarisoft Callbox BS comes with its own proprietary 5G Core software which is disabled in this setup, as a custom Open5Gs \cite{Open5GSproject} Core setup is used in the Server Cluster and MECs, to enable low-latency communication scenarios.

The edge computing node server cluster comprises of two identical servers with off-the-shelf parts that each contain a 26 Core, 56 Thread Intel Xeon Gold 6230R CPU @ 2.10GHz, 256 GB Error-Correcting Code (ECC) memory and a GeForce RTX 2080 SUPER GPU dedicated for AI/ML capabilities. 

For a cloud computing node, the Amazon Web Services (AWS) Paris node from the public cloud provider amazon was chosen as the closest cloud server to the site of evaluation in Barcelona, Spain. The robot, edge and cloud nodes were all a part of the same 5G network slice for the evaluation of the proposed NetROS-5G solution, as depicted in Fig. \ref{fig:slice}. Local, edge and cloud configurations were also studied. 

\subsection{Examined Scenario}

The effectiveness of the proposed framework NetROS-5G was assessed by conducting a use-case study involving an autonomous robot located in an airport, guiding the travelers to their designated departure gates.
The robot is scanning the faces of the travelers, matching it with the database entry and responding with a personalized message before navigating to the correct gate of the terminal.

\subsection{Examination Results}

\begin{figure}[t]
\centering
\includegraphics[width=\linewidth]{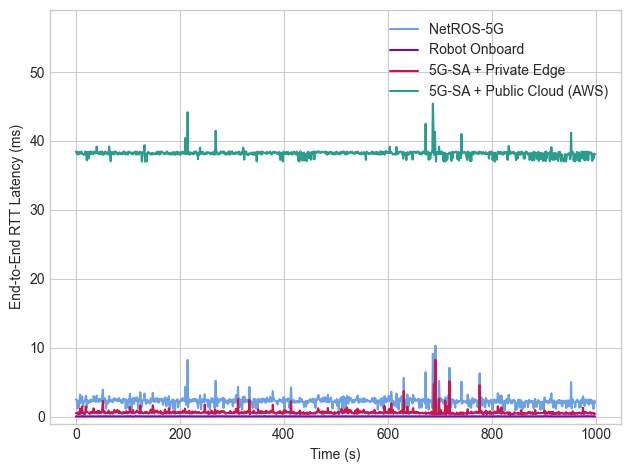}
\caption{End-to-end latency difference measured between the robot and the final destination, with 1000 samples.}
\label{fig:latplot}
\end{figure}

First, we examine the End-to-End (E2E) slice latency, between the robot and the furthest container dedicated to navigation, for all different settings. To test this Key Performance Indicator (KPI) we use teleoperation to measure the latency from the node that hosts the navigation container to the robot. This will allow measuring the network performance during remote operation. For this scenario, we take 1000 samples at 1Hz for all 4 different scenarios at the same time. As we can see in Fig. \ref{fig:latplot}, we observe that the 
overall latency when using only a cloud node for navigation can be as high as 199.832\% more than running it local at 38 ms average, compared to 0.016 ms average. The NetROS-5G framework is utilizing the Edge computing node for all navigation tasks, offloading data-related tasks on the cloud. As expected the response can be 183.548\% faster than this of the cloud for both settings.

Second, we examine the computing load yielded by the personalization functionality hosted in the nodes. For this KPI, we instantiate a Face-Recognition script, developed with Python and OpenCV, to showcase the computational load of a robot task that is used for personalization, recognizing the person that the robot interacts with. Table \ref{table:comp} shows the computing load ranges that were measured during the same 1000 samples. Using a cloud node for the personalization task uses minimal resources. Similarly, using an edge node as NetROS-5G to outsource the personalization task yields a lower load than performing the task locally in the onboard computer of the robot. It is apparent that a combination of edge and cloud computing is the best middle-ground between latency, computing capability, and cost.

\begin{table}[h!]
\centering
\begin{tabular}{|c|c|c|}
\hline
\textbf{Device} & \textbf{Robot CPU Usage} & \textbf{Memory Usage} \\ [0.5ex] 
\hline\hline
NetROS-5G & 12\%-18\% & 0.51 GB \\ 
\hline
Robot On-board & 65\%-80\% & 1.82 GB \\
\hline
5G-SA + Edge & 28\%-35\% & 0.95 GB \\
\hline
5G-SA + Cloud & 12\%-17\% & 0.51 GB \\
\hline
\end{tabular} \\ [1.5ex] 
\caption{Average computational load per scenario. 1000 Samples during test.}
\label{table:comp}
\end{table}

Finally, we examine the response time of the personalization task, measuring the time elapsed from the detection of a face that belongs to a traveler to the response in the local display. In order to perform this task in our laboratory environment with the simulated robot, we measured and compared the timestamps of the face detection and the response from the container that encapsulates the personalization app. According to the scenario, the robot is searching for the faces of the travelers, matching them to their photos in the database, and returning a personalized response as an action. Specifically, Table \ref{table:task} shows how prolonged the execution of the personalization task is locally on the onboard computer, almost 110.792\% slower than using the edge or the cloud node, making the networking delay insignificant for such tasks. Again, similar to the NetROS-5G, a fusion of the edge and cloud is the perfect middle-ground.

\begin{table}[h!]
\centering
\begin{tabular}{|c|c|c|}
\hline
\textbf{Device} & \textbf{Face recognition} & \textbf{Response time} \\ [0.5ex] 
\hline\hline
NetROS-5G & 657 sec & 698 sec \\ 
\hline
Robot On-board & 2354 ms & 2383 sec \\
\hline
5G-SA + Edge & 596 sec & 615 sec \\
\hline
5G-SA + Cloud & 587 sec & 684 sec \\
\hline
\end{tabular} \\ [1.5ex] 
\caption{Face recognition response time. 1000 Samples during test.}
\label{table:task}
\end{table}

\subsection{Discussion}

The performance of hardware and the reduction of network latency both contribute to the enhancement of personalization in Human-Robot Interaction (HRI) by enabling the robot to respond swiftly and precisely to inputs.
Low network latency is a critical aspect for real-time communication and task offloading in a remote host, as well as for the transmission of data used for personalization and analysis. It is a key factor in the efficient and accurate operation of real-time systems, as well as in the effective utilization of data for personalization and analysis. This highlights the importance of minimizing network latency in order to achieve optimal performance in real-time communication and the time required for data analysis and response.

\section{Conclusion}

In this paper, we have designed and developed a framework called NetROS-5G, to demonstrate the performance that can be exploited by utilizing network slicing in 5G networks, edge and cloud computing to overcome network latency and enhance the performance in personalization tasks in robotics.

As a next step in our research, we plan to conduct experiments utilizing Simultaneous Localization And Mapping (SLAM) navigation and planning techniques with a physical TurtleBot 4 robot, and further automating the NetROS-5G framework. The purpose of these additional experiments is to further evaluate the performance of the network and computational resources orchestration component.

\begin{acks}
This work was supported by the European research project 5G-ERA (101016681).
\end{acks}

\bibliographystyle{ACM-Reference-Format}
\bibliography{sample-base}

\end{document}